\documentclass[%
 reprint,
 amsmath,amssymb,
 aps,
]{revtex4-2}

\usepackage{xcolor}
\usepackage{graphicx}
\usepackage{dcolumn}
\usepackage{bm}

\DeclareMathOperator*{\argmin}{arg\,min}
\newcommand{\norm}[1]{\lVert#1\rVert_2}

\begin{document}

\preprint{APS/123-QED}

\title{Similarity-based equational inference in physics}

\author{Jordan Meadows$^{1}$}
 \email{jordan.meadows@postgrad.manchester.ac.uk}
\author{Andr\'e Freitas$^{1,2}$}%
\email{andre.freitas@manchester.ac.uk}
\affiliation{$^{1}$University of Manchester, Oxford Rd, Manchester, United Kingdom}
\affiliation{$^{2}$Idiap Research Institute, Rue Marconi 19, 1920 Martigny, Switzerland}%


\begin{abstract}
\indent Automating the derivation of published results is a challenge, in part due to the informal use of mathematics by physicists, compared to that of mathematicians. Following demand, we describe a method for converting informal hand-written derivations into datasets, and present an example dataset crafted from a contemporary result in condensed matter. We define an equation reconstruction task completed by rederiving an unknown intermediate equation posed as a state, taken from three consecutive equational states within a derivation. Derivation automation is achieved by applying string-based CAS-reliant actions to states, which mimic mathematical operations and induce state transitions. We implement a symbolic similarity-based heuristic search to solve the equation reconstruction task as an early step towards multi-hop equational inference in physics. 
\end{abstract}

\maketitle


\section{Introduction}

\indent Automating physical reasoning first involves the comprehension of physical concepts, language, and algebra, which form a cohesive informal mathematical explanation. Physics-inspired data-driven neural approaches are used often for the purpose of accurate calculation and simulation, but not for deriving equations~\cite{kissas2020machine, liang2019phillips, teichert2019machine, collins2018anomaly, bereau2018non, torlai2018neural}. Examples of equational~\cite{udrescu2020symbolic, udrescu2020ai, kim2020integration} and conceptual~\cite{iten2020discovering} inference do not convey the complex arguments conducted within derivations, which combine assumed mathematics similar to premise selection, symbolic manipulation of equations, and reference to physical concepts. Automated Theorem Proving (ATP) is associated with mathematical rigour, but is not suitable for this type of informal equational and conceptual reasoning used by physicists~\cite{davis2019proof, kaliszyk2015formalizing}. Literature at this interface is limited and scarce~\cite{govindarajalulu2015proof}, and there are few complete real-world derivations of published results that exist in a computer interpretable format. \newline \indent Considering these limitations, this paper proposes three contributions: (i) we present a novel dataset consisting of 368 equations compiled from the detailed derivation of a published equation in physics~\cite{mann2018manipulating}, represented as a sequence of states and actions, and we describe the dataset creation method; (ii) we define an \textit{equation reconstruction task} by considering the smallest non-trivial derivation as three consecutive equational states with an unknown intermediate state we aim to reconstruct; (iii) we propose the use of \textit{symbolic similarity-based heuristics}, a knowledge base of assumed equations, and a set of granular mathematical operations posed as string-related actions formulated in a computer algebra system (CAS), to solve the equation reconstruction task on the PhysAI-DS1 dataset~\footnote{https://github.com/jmeadows17/equational-inference}. We claim an approach accuracy of 56.2\% considering both exact matching and the non-zero similarity cutoff regime, given the dataset, knowledge base, the set of actions, and the computer algebra system. 

\section{Equational Inference in Physics}

\indent Research at the border between artificial intelligence and physics has seen many recent successes. Data-driven approaches are popular, where many share the theme of feeding large datasets to physics-inspired architectures, learning physical model parameters often within the context of real-world problems~\cite{kissas2020machine, liang2019phillips, teichert2019machine,  wu2018physics, deringer2017machine, ramakrishnan2015big, baldi2014searching}. In contrast to these simulation and calculation themed approaches, others aim to infer equations or conceptual insights from experimental or generated data from toy model physics closer to symbolic regression~\cite{iten2020discovering, udrescu2020ai, udrescu2020symbolic,  wu2019toward, kim2020integration, raissi2018hidden}, even recovering the latent network structure of dynamical systems from time series data \cite{zhang2019general}. While this class of approach is more similar to traditional theoretical physics, neural methods suffer from an explainability problem, and equational and conceptual reasoning are crucial components of physics derivations. \newline \indent Automated theorem proving has seen little success in physics on the equational reasoning side, literature is scarce~\cite{govindarajalulu2015proof}, and efforts have been made to formalise physics towards ATP~\cite{kaliszyk2015formalizing}. While Ref.~\cite{davis2019proof} highlights five directions, we expand upon three in this work: (i) increasing the collection of physical theories we have in forms that can be used for training and validating symbolic reasoners in physics; (ii) developing representation methods and evaluation tasks which necessitate approaches that select suitable approximations, idealisations and abstractions; and (iii) analysis of the nature of the informal argumentation used in physics. \newline \indent Aligned with direction (i), we present 368 equations taken from physical theory~\cite{mann2018manipulating} converted into a dataset as the first contribution. Following direction (iii), this dataset is generated by a method which captures much of the derivation argumentation, by treating derivation progression as a finite state machine reliant upon computer algebra operations and representations. Direction (ii) we address with an \textit{equation reconstruction task} as a second contribution, which provides a medium for creating approaches which involve the selection of idealisations, approximations, and abstractions. As a third contribution, our similarity-based approach partially reconstructs the equational argument from PhysAI-DS1, which requires use of symbolic approximations and the selection of supporting premises as a result of the physical theory the comprised derivation represents. \newline \indent ATP methods have a strong reliance on logical formalisation. In order to be comparable to ATP-based methods, derivations in physics would need to be translated into a logical form. There is a shared understanding that existing standard logical frameworks used in ATPs are limited and not aligned to the requirements of the nature of physics argumentation \cite{kaliszyk2015formalizing, davis2019proof}, and thus comparative baselines of this kind can not be made apart from a very controlled fragment of the physics domain. Additionally, to our knowledge there are no efforts automating computer algebra in physics, not at least at the discourse-level form expressed in physics papers. We claim our baseline for the equation reconstruction task is the first of its kind, with other approaches unable to be easily adapted to solve it.\newline \indent Reinforcement learning has been applied in the context of regular~\cite{kaliszyk2018reinforcement} and equational~\cite{piepenbrock2021learning} theorem proving, and separately nuclear physics~\cite{luo2018automatic}. Although such techniques may be useful for searching large state spaces, they are not currently applied in the context of computer algebra expressed at the surface form of physics papers. Instead, our contribution targets a granular understanding of the dialog between equational symbolic forms in physics and similarity metrics, and its utility as one component of the symbolic inference mechanism. \newline \indent We develop and evaluate our early single-hop~\cite{min2019compositional} approach in the context of a multi-hop extensible equational inference task, using a computer algebra sequence and subsequent dataset created using a novel method for converting contemporary physics derivations into CAS-interpretable data.

\begin{figure*}[htp!]
    \centering
    \includegraphics[width=1.5\columnwidth,
	height=1\columnwidth]{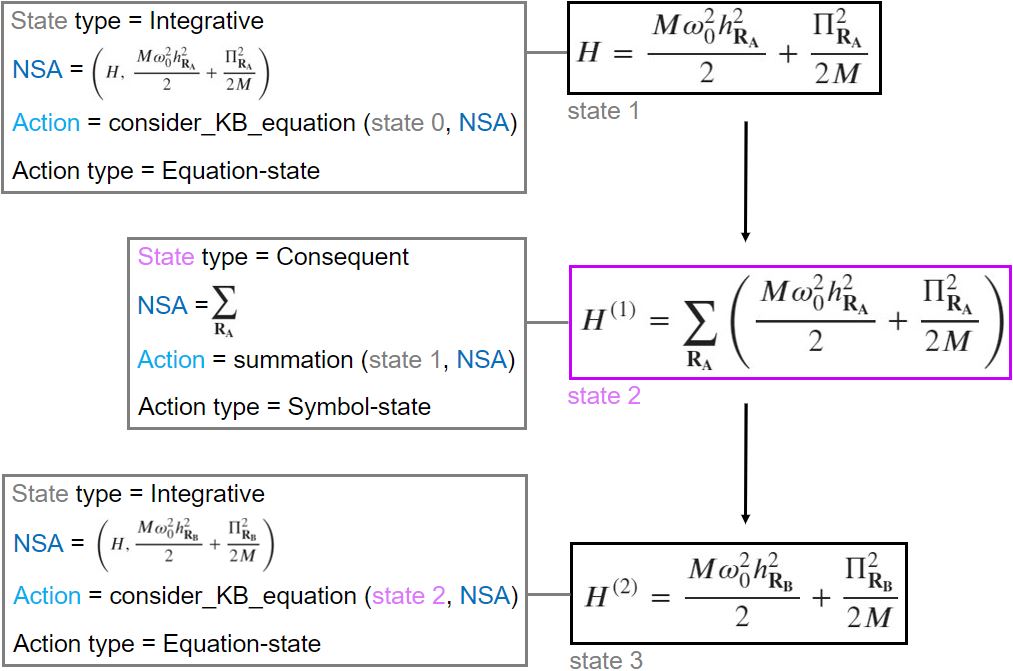}
    \caption{The first three consecutive equational states from the computer algebra derivation. Each state is seen associated with four main pieces of information stored in PhysAI-DS1. If \textit{state 2} is considered unknown, the three states form a \textit{derivation unit} and this diagram represents the equation reconstruction task, where the goal is to select the correct action and NSA to reconstruct \textit{state 2} given states 1 and 3.}
    \label{fig:my_label}
\end{figure*}

\section{Dataset Construction Methodology}
 
\indent In theoretical condensed matter, Ref.~\cite{mann2018manipulating} is concerned with manipulating topological features of polaritons in cavity-embedded honeycomb metasurfaces. The following equation is the first numbered result:

\begin{eqnarray}
H_{mat} =\text{\,}&& \hbar \tilde{\omega}_0 \sum_{\textbf{q}}\Big(a^{\dagger}_\textbf{q}a_\textbf{q} + b^{\dagger}_\textbf{q}b_\textbf{q}\Big)\nonumber\\
&&\text{\,}+ \hbar\tilde{\Omega}\sum_{\textbf{q}}\Big(f_{\textbf{q}}b^{\dagger}_{\textbf{q}}a_{\textbf{q}} + H.c\Big).
\end{eqnarray}

\indent \newline This is accompanied by a brief derivation highlighting key equations. The challenge is to use this derivation as a guide to derive (1) with \textit{CAS-level representation}, at the granularity level of physicists performing the derivation in real time, to then store this derivation as a dataset. 

\subsection{Manual Derivation} \indent We start by manually deriving (1) from a single first-quantised harmonic oscillator (as seen in Figure 1). We avoid skipping any steps, which drastically lengthens the derivation, as in practice minute mathematical operations are taken for granted at the paper form, which may require its elicitation to enable CAS-compliance. We frame the derivation as a finite state machine where the current equation is the current state, and we carefully choose an informal action (operation), and action argument, that will progress the derivation to the next state. The state, action, and argument sequences are recorded up to equation (1). 

\subsection{Computer Algebra Augmentation} \indent This curated derivation serves as a guide for the approximate derivation to be recreated in the CAS. However, particular actions, e.g. ``divide RHS by 2", may not already exist explicitly in the CAS and must be constructed. There are limitations on how actions can be constructed, so the reconstructed derivation may not exactly mirror the hand-written version in general. \newline \indent We use SymPy~\cite{meurer2017sympy}, which allows the rendering and manipulation of equations in \LaTeX{}, such that equations can match those within papers. From the hand-written derivation which we represent as a \textit{state sequence}, we define an initial 2-tuple (LHS, RHS) equational state using LHS and RHS generated from the CAS, and build and store appropriate actions which induce state transitions to other 2-tuples, which closely match the hand-written state sequence where possible. Along the way, actions may require yet unused symbols and assumed equations to progress the derivation, both of which we collect in a \textit{knowledge base (KB)}. We employ the following Conventions:

\begin{enumerate}
    \item Equations are stored as (LHS, RHS) tuples.
    
    \item Actions are functions which accept \textit{two arguments} including the current state and: no argument, a symbol, an equation. These secondary arguments are called Non-State Arguments (NSAs). Respectively these action categories are called \textbf{self-state}, \textbf{symbol-state}, and \textbf{equation-state} actions. 
    
    \item Actions cause the \textit{minimum possible state transition} while still progressing down the derivation sequence. In contrast, the action causing the maximum possible state transition would be called ``immediately derive the goal state from the initial state". A more granular action example is ``divide RHS by 2" which induces a minimal change in the state semantics comparatively. Actions are designed to operate at this resolution, but physicists work less granularly.
    
    \item No LHS is the same for any state in the sequence (by default). For example, if $(H, \hbar \omega_0)$ is the current state and the action ``divide RHS by 2" is applied to it, then the next state in the sequence is $(H^{(1)}, \hbar \omega_0/2)$ with an index appended to the LHS, because $\hbar \omega \neq \hbar \omega/2$. However an action ``remove index" exists for the specific case of algebraic manipulation with the LHS.
    
    \item If an action is not appropriate given its arguments, then the action returns the current state. \newline (e.g. action = ``divide current state RHS by KB equation RHS", but the KB equation RHS is a vector). 
    
\end{enumerate}

\indent From this sequence of states, actions, and arguments, we construct a dataset \textbf{PhysAI-DS1}, consisting of 368 consecutive entries. Features from PhysAI-DS1 from left to right read: \textit{\{State string length (LaTeX text), State string (LaTeX text), State string length (SymPy text), State string (SymPy text), State string length (SymPy tree), State string (SymPy tree), Action, Non-state argument (SymPy text), State type, Action type\}}, which we have designed to capture the important aspects of a physics derivation. The minimum, average, and maximum equation string lengths for the SymPy text considered in this paper are respectively 52, 495, 5476. The \LaTeX{} equivalent is 56, 582, 6318. We consider three different representations for the states, while NSAs are expressed using the native representation of the CAS. Of the 368 states, 76 are categorised as type \textit{integrative}, 217 are \textit{consequent}, and 75 are \textit{terminal}. \newline \indent The states are categorised by their role in the derivation: \textbf{Integrative states:} formed by the action ``consider knowledge base equation".
\textbf{Consequent states:} follow integrative states in the sequence. \textbf{Terminal states:} directly precede integrative states. There are 35 unique actions split into self-state, symbol-state and equation-state categories, described in Convention 2. The final state 2-tuple in the dataset represented as an equation is given by

\begin{eqnarray}
H^{(83)}=&&\text{\:}\hbar\sum_{\textbf{q}}\tilde{\Omega}\Big(H.c + f_{\textbf{q}}b^{\dagger}_{\textbf{q}}a_{\textbf{q}}\Big)\nonumber\\
    &&\text{\,}+ \hbar \sum_{\textbf{q}}\tilde{\omega}_0\Big(a^{\dagger}_\textbf{q}a_\textbf{q} + b^{\dagger}_\textbf{q}b_\textbf{q}\Big)
\end{eqnarray}

\indent which is mathematically equivalent but rearranged from (1), up to the LHS. SymPy has a native method of ordering equations and this ordering problem persists throughout the derivation. The order of non-commutative terms is preserved however, and this issue is cosmetic. The names of actions are in some cases misleading and indeed behave unintuitively also. This reflects the limitations of our chosen CAS, as often the most obvious way to force a desired state transition during construction involves the invention of obscure string-based operational actions, as a result of the limitations of the computer algebra system and the curated derivation. See Figure 1 for key information associated with equational states in PhysAI-DS1.

\section{Equation Reconstruction Task}

\indent The \textit{equation reconstruction task} aims to support the construction of AI-supported derivation systems operating at different levels of symbolic representation (from a string/natural language level to an algebraic object/operational level). The task considers three consecutive states in a derivation grouped into a \textit{derivation unit} $\{s_{i-1}, s_i, s_{i+1}\}$, where two consecutive actions performed on $s_{i-1}$ form the unit. \newline \newline \indent \textit{We aim to create a reconstruction $\hat{s}_i$ such that it is mathematically equivalent and sufficiently similar to an unknown $s_i$, given $s_{i-1}$ and $s_{i+1}$}. \newline \newline \indent In practice this involves selecting the correct action and NSA to cause the transition $s_{i-1} \rightarrow s_i$ (with probability 1) given a suitable knowledge base and action set. This derivation unit translates along the state sequence, and combined with a suitable inference algorithm, outputs one reconstruction per translation. \newline \indent States are non-Markov which reflects the behaviour of physicists who may use equations in the distant history to progress the derivation. Action and state sequences behave deterministically, and actions receive two arguments: the current state and the non-state argument (NSA). In addition to those in the KB, we allow for equation-type NSAs to be any equational state in the history up to $s_i$ which violates the Markov condition. \newline \indent A derivation unit treats the contained states as a ``micro-derivation" with no explicit connection to the complete sequence. Although the state history is required by the actions for progression, it effectively becomes part of the KB which grows proportional to $i$. As follows, for $N$ states in the full derivation, there are $N-2$ fully independent reconstructions $\hat{s}_i$, which is equivalent to solving the equation reconstruction task for $N-2$ independent derivations, each of size three. \newline \indent We append dummy states onto the end and beginning of the sequence totalling $N = 370$ states. This results in 368 reconstructions comparable with the 368 states in the derivation. The first dummy state is given by $s_0 = (x, ?)$ and represents an entirely neutral equation akin to beginning with no prior knowledge. The second dummy state appended last is the 2-tuple form of equation (1) from Ref.~\cite{mann2018manipulating} \textit{passed through a computer algebra system}, such that the final state in PhysAI-DS1, and the dummy state, are given respectively as:

\begin{eqnarray}
s_{N-1} =&&\text{\,} \bigg(H^{(83)}, \:\hbar\sum_{\textbf{q}}\tilde{\Omega}\big(H.c + f_{\textbf{q}}b^{\dagger}_{\textbf{q}}a_{\textbf{q}}\big)\nonumber\\
    &&\text{\,} + \hbar \sum_{\textbf{q}}\tilde{\omega}_0\big(a^{\dagger}_\textbf{q}a_\textbf{q} + b^{\dagger}_\textbf{q}b_\textbf{q}\big)\bigg)
\end{eqnarray}

\begin{eqnarray}
s_{N} =&&\text{\,} \bigg(H_{mat}, \:\hbar\sum_{\textbf{q}}\tilde{\Omega}\big(H.c + f_{\textbf{q}}b^{\dagger}_{\textbf{q}}a_{\textbf{q}}\big)\nonumber\\
    &&\text{\,} + \hbar \sum_{\textbf{q}}\tilde{\omega}_0\big(a^{\dagger}_\textbf{q}a_\textbf{q} + b^{\dagger}_\textbf{q}b_\textbf{q}\big)\bigg).
\end{eqnarray}

\section{Symbolic Similarity-based Search}

\subsection{Similarity Measures}

\indent If we consider equations (3) and (4) in the sequence we can observe they are equivalent but symbolically dissimilar, as the string representations of each LHS are not identical. We employ a similarity measure $M$ to classify whether one state is equal \textit{and similar} to another, with a hyperparameter $\varepsilon$. \newline \indent We form a set of canonical algebraic actions referred by the PhysAI-DS1 dataset. Actions accept two arguments: the current state, and a Non-State Argument (NSA). An NSA can be either None, a symbol, or an equation 2-tuple.
\newline \indent We form a knowledge base of requisite equations necessary for the derivation, and a set $L$ consisting of every (non-numeric) symbol from each equation. All states prior to $s_{i-1}$ are also included in the knowledge base. Additionally, if $l_{i-1}$ is the set of symbols which constitutes $s_{i-1}$, and $l_{i+1}$ is the set which constitutes $s_{i+1}$, then we append the symmetric difference $l_{i-1}\, \triangle \, l_{i+1}$ to $L$. As we know the form of $s_{i-1}$ and $s_{i+1}$, we can use the constituent symbols to guide the derivation which we achieve by including $l_{i-1}\, \triangle \, l_{i+1}$ in $L$, an element of which may be accepted as the argument of an action function as an NSA. The set $R$ contains the state history up to $s_i$ and the requisite equations of the knowledge base. We define a combined knowledge base $\mathcal{K} = L \cup R$. 
\newline \indent Given the knowledge base $\mathcal{K}$, the set of actions $\mathcal{A}$, and the initial and goal states respectively $s_{i-1}, s_{i+1} \in \mathcal{S}$ for set of states $\mathcal{S}$, we aim to generate a state $\hat{s}_i \in \mathcal{S}$ such that the similarity measure $M(\hat{s}_i, s_i) \leq \varepsilon$ where $M: \mathcal{S} \times \mathcal{S} \rightarrow \mathbb{R}_{\geq 0}$, and $\varepsilon \in \mathbb{R}_{\geq 0}$ is a hyperparameter determining whether $\hat{s}_i$ and $s_i$ are considered both equal and similar. We choose $M$ to be based upon one of five similarity measures:
\begin{itemize}
    \item Levenshtein distance, $M(s_i, s_j) = M_L(s_i, s_j)$: Minimum number of single character insertions, substitutions or deletions required to transform string $s_i$ into $s_j$.
    
    \item Damerau-Levenshtein distance, $M(s_i, s_j) = M_{DL}(s_i, s_j)$: Extension of Levenshtein distance to include single character transpositions.
    
    \item Hamming distance, $M(s_i, s_j) = M_H(s_i, s_j)$: Minimum number of single character substitutions to transform $s_i$ into $s_j$.
    
    \item Jaro similarity, $M(s_i, s_j) = 1 - \text{sim}_J(s_i, s_j)$: \begin{equation*} \resizebox{0.42\textwidth}{!}{$\text{sim}_J(s_i, s_j) = 
        \begin{cases}
        0 & \text{if } m = 0\\
        \frac{1}{3}\Big(\frac{m}{|s_i|}+\frac{m}{|s_j|} + \frac{m-t}{m}\Big) & \text{otherwise}
        \end{cases}$}
    \end{equation*} where $|s_i|$ is string length, $m$ is the number of matching characters (considered matching if \newline $\lfloor\text{max}(|s_i|,|s_j|)/2\rfloor - 1$ or less characters between them), and $t$ is half of the number of transpositions.
    
    \item Jaro-Winkler similarity, $M(s_i, s_j) = 1 - \text{sim}_{JW}(s_i, s_j)$: \begin{equation*}
        \text{sim}_{JW} = \text{sim}_{J} - lp(1 - \text{sim}_J)
    \end{equation*} where $l$ is the length of a common prefix at the start of the string up to a maximum of 4 characters, and $p$ is a constant scaling factor (defined by the Jellyfish library default for our case).
    
\end{itemize} Taking $M$ as the Levenshtein measure, the distance between $s_N$ and $s_{N-1}$ from the derivation state sequence is $M(s_{N-1}, s_N) = 5$, therefore we require $\varepsilon \geq 5$ to determine the two states are similar within the Levenshtein metric. Lower bounds on $\varepsilon$ that capture the final reconstruction for each of the similarity measures can be obtained via \begin{equation}
    \varepsilon_{low} = M(s_N, s_{N-1}).
\end{equation}

\subsection{Heuristic Search}

\indent Following the \textit{equation reconstruction} task description, while we know the form of $s_{i-1}$ and $s_{i+1}$ with the $s_i$ unknown, we can form reconstruction \textit{candidates} $\hat{c}_i$ by applying an action $a \in \mathcal{A}$ to $s_{i-1}$ such that $a:(s_{i-1}, k) \mapsto \hat{c}_i$, where $\hat{c}_i \in \mathcal{S}$ and $k \in \mathcal{K}$. We apply a second action to all possible $\hat{c}_i$ with the mapping $a^\prime:(\hat{c}_i, k^\prime) \mapsto \hat{c}_{i+1}$, where in general $k \neq k^\prime$ and $a \neq a^\prime$, to obtain $\hat{c}_{i+1}$ for $(\hat{c}_i, \hat{c}_{i+1}) \in \mathcal{S}_{final}$. \newline \indent At this stage multiple paths $\{s_{i-1}, \hat{c}_i, \hat{c}_{i+1}\}$ exist. By solving the following optimisation problem, we can determine $\hat{s}_{i+1}$:

\begin{equation}
\hat{s}_{i+1} = \argmin_{(\hat{c}_i, \hat{c}_{i+1}) \in \, \mathcal{S}_{final}}\,M({\hat{c}_{i+1}, s_{i+1}}).
\end{equation}

\indent \newline \newline We employ use of a heuristic $H(\hat{c}_i, s_{i+1}):\mathcal{S}\times \mathcal{S} \rightarrow \mathbb{R}_{\geq 0}$ to order the paths, as the search may end early under a specific condition. If we compare a state $\hat{c}_{i+1}$ such that $M(\hat{c}_{i+1}, s_{i+1}) = M(\hat{s}_{i+1}, s_{i+1}) = 0$, then the reconstruction candidate $\hat{c}_i$ from the path containing $\hat{c}_{i+1} = \hat{s}_{i+1}$ is selected as the reconstruction $\hat{s}_i$ which ends that iteration. Otherwise, all $\hat{c}_{i+1}$ are compared from each path, and the $\hat{c}_{i}$ of the path corresponding with the lowest $M(\hat{c}_{i+1}, s_{i+1})$ is taken as the $\hat{s}_i$. Based on the Levenshtein metric, $H$ is defined as follows:

\begin{itemize}
    \item $x(\hat{c}_i, s_{i+1}) = \lvert l_s \setminus l_c\rvert$, where $l_s$ and $l_c$ are the set of symbols that constitute $s_{i+1}$ and $\hat{c}_i$ respectively. If $x$ is large that means there are many symbols in $s_{i+1}$ that are not in $\hat{c}_i$, and the two are unlikely to be semantically linked. 
    
    \item $y(\hat{c}_i, s_{i+1}) = M_L(c_{min}, s_{min})$, where $c_{min}$ and $s_{min}$ are the \textit{subexpressions} in the trees of $\hat{c}_i$ and $s_{i+1}$ respectively corresponding to the lowest $M_L$, given that the number of characters in the string representations of $c_{min}$ and $s_{min}$ is \textit{less than or equal to 100}. This represents a similarity calculation for \textit{small} subexpressions. If no suitable comparisons exist then $y = M_L(\hat{c}_i, s_{i+1})$.
    
    \item $z(\hat{c}_i, s_{i+1}) = M_L(c_{max}, s_{max})$, where $c_{max}$ and $s_{max}$ are the \textit{subexpressions} in the trees of $\hat{c}_i$ and $s_{i+1}$ respectively corresponding to the lowest $M_L$, given that the number of characters in the string representations of $c_{min}$ and $s_{min}$ is \textit{greater than 100}. This represents a similarity calculation for \textit{large} subexpressions. If no suitable comparisons exist then $z = M_L(\hat{c}_i, s_{i+1})$.
\end{itemize}

\noindent Given $\textbf{r} = \big(n_1x(\hat{c}_i, s_{i+1}), n_2y(\hat{c}_i, s_{i+1}), n_3z(\hat{c}_i, s_{i+1})\big)$ \newline for $n_1, n_2, n_3 \in \mathbb{Z}_{\geq 0}$, the heuristic is given by

\begin{equation}
    H(\hat{c}_i, s_{i+1}; n_1, n_2, n_3) = \norm{\textbf{r}(\hat{c}_i, s_{i+1}; n_1, n_2, n_3)}^2.
\end{equation}

\indent \newline Each path $\{s_{i-1}, \hat{c}_i, \hat{c}_{i+1}\}$ corresponds to a heuristic value and lowest-valued paths are searched first with the aim of finding a comparison $M(\hat{c}_{i+1}, s_{i+1}) = 0$ which terminates the search, else all comparisons are made. \newline \indent The process is based on the notion that \textit{if $s_{i+1}$ can be obtained by applying two consecutive actions to $s_{i-1}$, then the state after one action may be $s_i$.} \newline \indent Extending to the multi-hop case, for number of intermediate states $\mathcal{N}$, the number of derivation paths $\{s_{i-1}, c_i, ..., c_{i+\mathcal{N}}\}$ grows as $(|\mathcal{A}||\mathcal{K}|)^{\mathcal{N} + 1}$. Then, each Levenshtein comparison $M_{L}(\hat{c}_{i+\mathcal{N}-1}, s_{i+\mathcal{N}})$ in the heuristic, and final \textit{e.g.} Damerau-Levenshtein comparison in the search $M_{DL}({\hat{c}_{i+\mathcal{N}}, s_{i+\mathcal{N}}})$ has $O(mn)$ time complexity for equation string lengths $m$ and $n$. However, this work targets step-wise, single-hop inference~\cite{min2019compositional} as a unit of analysis, and end-to-end or multi-hop inference is currently outside our scope.

\section{Empirical Evaluation}

\indent A reconstruction $\hat{s}_i$ is classified as a success in the evaluation if $M(\hat{s}_i, s_i) \leq \varepsilon$, where $\varepsilon = \eta\varepsilon_{low}$ for $\eta \in \mathbb{Z}_{\geq 0}$. The similarity value $\varepsilon_{low} = M(s_{N-1}, s_N)$ for state sequence of length $N$, is necessary in order to classify equations (3) and (4) as equal and similar after a second action, and represents a unit of difference scalable with $\eta$. It can be found directly by comparing the penultimate and final equations (states) in the derivation (state sequence). The value $\eta = 0$ represents exact string matching, and $\eta = 1$ represents the minimum similarity required to reconstruct all equational states. At $\eta = 0$ false negatives may occur with no false positives and the accuracy is minimised, while for $\eta > 0$ the accuracy increases due to decrease in false negatives and increase in false positives. We evaluate for accuracy at values of $\eta \in \{0,1\}$. Table 1 describes the approach accuracy from the equation reconstruction experiments across the similarity measures.
{\renewcommand{\arraystretch}{1.3}
\begin{table*}[htp!]
\label{sample-table}
\begin{center}
\begin{tabular}{c c c c}
\hline
Similarity Measure & $\varepsilon_{low}$ & Accuracy $(\eta = 0)$ & Accuracy $(\eta=1)$\\
\hline
	        Levenshtein & 5 & 0.516 & 0.533\\
			Damerau-Levenshtein & 5 & \textbf{0.562} & 0.562\\ 
			Hamming & 224 & 0.508 & \textbf{0.799}\\
			Jaro & 0.1643 & 0.514 & 0.690\\
			Jaro-Winkler & 0.0986 & 0.514 & 0.690\\
			\hline
\end{tabular}
\caption{The approach accuracy from the equation reconstruction experiments across the similarity measures. $\varepsilon_{low} = M(s_{N-1}, s_N)$ can be found directly by comparing the penultimate and final equations (states) in the derivation (state sequence). The hyperparameters used for each experiment were $(\varepsilon, n_1, n_2, n_3) = (\varepsilon_{low}, 10, 10, 10)$, where $\{n_i\}$ refer to the hyperparameters associated with equation (7), the heuristic.}
\end{center}
\end{table*}}

\indent From the Table 1 results, the approach based on the Damerau-Levenshtein metric outperforms the other string measures with respect to exact matches, and does not differ in accuracy upon introducing a unit of difference $\varepsilon_{low}$. This suggests that Damerau-Levenshtein is less susceptible to the inclusion of false positives with increasing $\eta$. The Levenshtein-based measures both show low sensitivity to increasing $\eta$.\newline \newline \indent The Hamming measure approach results in the lowest exact matching accuracy, but the highest accuracy at unit difference $\varepsilon_{low}$. This suggests that the measure is relatively more susceptible to false positives within our formalism, and less suitable for this task.

\subsection{Categories of States and Actions}

\indent Actions may be categorised dependent upon their NSA as either \textit{self-state}, \textit{symbol-state}, or \textit{equation-state}. States are categorised by their role in the state sequence as either \textit{integrative}, \textit{consequent}, or \textit{terminal} (see Section 3). There are two actions per state reconstruction (see Section 4) but only one action is responsible for the reconstruction directly. We consider this action, and the corresponding state, in the following analysis. Table 2 describes the total number of cases for each category of state versus each category of action. Table 3 describes the accuracy corresponding to each case. \newline

{\renewcommand{\arraystretch}{1.3}
\begin{table*}[htp!]
\label{sample-table}
\begin{center}
\begin{tabular}{l  c  c  c}
			\hline
			& \multicolumn{3}{c}{Action Type (Count)}\\
			State Type & \multicolumn{3}{c}{||||||||||||||||||}\\
			& Self-state & Symbol-state & Equation-state\\
			\hline
			Integrative & 0 & 0 & 76\\
			Consequent & 86 & 31 & 100\\
			Terminal & 20 & 18 & 37\\
			\hline
		\end{tabular}\\
	\caption{Number of instances associated with combinations of action type and state type pairs.}
	\end{center}
\end{table*}}

{\renewcommand{\arraystretch}{1.3}
\begin{table*}[htp!]
\label{sample-table}
\begin{center}
\setlength{\tabcolsep}{5pt}
	\resizebox{\textwidth}{!}{%
    \begin{tabular}{c  c  c c c  c c c}
    			\hline
    			& & \multicolumn{3}{c}{Action Type Accuracy $(\eta = 0)$} & \multicolumn{3}{c}{Action Type Accuracy $(\eta = 1)$}\\
    			Similarity Measure & State Type & \multicolumn{3}{c}{||||||||||||||||||} & \multicolumn{3}{c}{||||||||||||||||||}\\
    			& & Self-state & Symbol-state & Equation-state & Self-state & Symbol-state & Equation-state\\
    			\hline
    			Levenshtein & Integrative & - & - & 0.776 & - & - & 0.829\\
    			& Consequent & 0.605 & \textbf{0.742} & 0.560 & 0.628 & 0.742 & 0.560\\
    			& Terminal & 0.000 & 0.000 & 0.000 & 0.000 & 0.000 & 0.000\\
    			\hline
    			Damerau-Levenshtein & Integrative & - & - & \textbf{0.868} & - & - & 0.868\\
    			& Consequent & \textbf{0.663} & 0.710 & \textbf{0.620} & 0.663 & 0.710 & 0.620\\
    			& Terminal & 0.000 & 0.000 & 0.000 & 0.000 & 0.000 & 0.000\\
    			\hline
    			Hamming & Integrative & - & - & 0.776 & - & - & 0.921\\
    			& Consequent & 0.593 & 0.710 & 0.550 & \textbf{0.919} & \textbf{1.000} & 0.760\\
    			& Terminal & 0.000 & 0.000 & 0.000 & \textbf{0.750} & \textbf{0.556} & \textbf{0.351}\\
    			\hline
    			Jaro & Integrative & - & - & 0.776 & - & - & \textbf{0.961}\\
    			& Consequent & 0.605 & 0.710 & 0.560 & 0.767 & 0.871 & \textbf{0.880}\\
    			& Terminal & 0.000 & 0.000 & 0.000 & 0.000 & 0.000 & 0.000\\
    			\hline
    			Jaro-Winkler & Integrative & - & - & 0.776 & - & - & \textbf{0.961}\\
    			& Consequent & 0.605 & 0.710 & 0.560 & 0.767 & 0.871 & \textbf{0.880}\\
    			& Terminal & 0.000 & 0.000 & 0.000 & 0.000 & 0.000 & 0.000\\
    			\hline
    		\end{tabular}
    		}
    	\caption{Equation reconstruction accuracy associated with combinations of action type and state type pairs. Bold numbers represent highest values from the seven possible combinations, across the five approaches.}
    	\end{center}
\end{table*}}
	
	\indent At $\eta = 0$, consequent states formed by self-state actions and equation-state actions, and all integrative states, are most accurately reconstructed by the Damerau-Levenshtein approach. The standard Levenshtein approach reconstructs consequent states with symbol-state actions most accurately. Even in the exact matching case at $\eta = 0$, \textit{integrative states are reconstructed with almost 90\% accuracy, and all terminal states fail.} \newline \newline \indent At $\eta = 1$ the Levenshtein-based approaches share the lowest performance, where the Hamming approach displays non-zero accuracy in terminal states, and over 0.9 accuracy in the consequent self-state and symbol-state categories. This suggests the Hamming approach quickly includes false positives with increasing $\eta$, as do the Jaro-based approaches particularly within the equation-state class, comparative to the Levenshtein-based approaches.
	
\subsection{Qualitative Analysis and Approach Limitations}
 
 \indent To provide additional insights into the limitations of our approach to solving the equation reconstruction task, we categorise known reconstruction failure types for the Damerau-Levenshtein variant. \newline \newline \indent \textbf{Terminal State Problem.} All terminal state reconstructions will fail - a claim supported by Table 3. This arises due to the state after being \textit{integrative} by definition, and integrative states are reconstructed by the specific action ``consider knowledge base equation". Integrative states represent the beginning of new derivation branches, and arriving at an integrative state after two consecutive actions is only dependent upon the second action being ``consider knowledge base equation". Therefore, any possible first action and subsequent reconstruction is suitable because of this condition on the second action. Terminal states represent 20.4\% of the test set and around half of all reconstruction errors. \newline \indent \textbf{Multiple Pathways.} Reconstructions are ultimately determined by the final similarity check $M(\hat{c}_{i+1}, s_{i+1})$ such that $\hat{c}_{i+1}$ satisfies equation (6). Many $\hat{c}_{i}$ may be associated with a unique $\hat{c}_{i+1}$, and the lowest heuristic value $\hat{c}_{i}$ is then selected. \newline \indent \textbf{Repeating equations.} If the $s_{i+1}$ is an equation that has already been derived in the state history (up to the LHS index) then the first action will be ``consider knowledge base equation" with the NSA as the repeated equation, and a second action chosen to not transition the state (equivalent to ``do nothing", Convention 5). \newline \indent \textbf{Action Limitations.} Actions are made specifically to induce state transitions between \textit{known} states, \textit{i.e.} from the curated derivation passed through CAS. This means upscaling the data or considering other datasets is time-consuming, as the set of consecutive actions that form a separate derivation may be entirely different. This would mean for each new derivation at least one pass of the full derivation sequence through CAS must be made, and a new dataset must be constructed. Additionally, even within the same derivation, actions may not return mathematically correct states in all cases, as many are reliant on string manipulation and are not native to SymPy directly. There is considerable advantage in finding mathematically ``robust" actions suitable for rigorous exploration of unknown states, but such work is outside our current scope. \newline \indent \textbf{Knowledge Base Selection.} Premise selection is the task of selecting relevant mathematical statements aiming to maximise the probability of proving a given conjecture. The set of knowledge base equations have been hand-crafted in the same vein as the action set, and they suffer similar limitations. It is a non-trivial task~\cite{ferreira-freitas-2020-premise} selecting suitable axiomatic equations or supporting statements capable of reconstructing derivations or proofs, and is outside our current scope. \newpage

\section{Conclusion}

\indent In response to demand for informal mathematical datasets~\cite{davis2019proof}, we have created a method for producing datasets from physics derivations as a step towards multi-hop inference in physics. This method first involves a curated hand-written derivation, then a recreation of the derivation in a suitably expressive computer algebra system, under an expressive representation scheme. We categorise states and actions within this methodology, and employ it to create a dataset \textbf{PhysAI-DS1} from a recent result in condensed matter~\cite{mann2018manipulating}. We propose an equation reconstruction task which considers sets of three consecutive states within the derivation. The intermediate state is then considered unknown, and information from a knowledge base and the initial and final states may be used to mathematically infer the form of the unknown state by applying sequences of string-based CAS-reliant actions to the known state, and by utilising similarity measures to approximate equality. As an early solution to the equation reconstruction task, we formulate primitive similarity-based heuristics with a search algorithm capable of reconstructing equations within PhysAI-DS1 to 56.2\% accuracy using the Damerau-Levenshtein metric, compared to baseline approaches using four distinct string similarity measures. We discuss limitations of the CAS-reliant similarity-based approach, and systematically characterise the performance of separate categories of action and state. Future work will focus on the physics explanation extraction process to account for different physical systems in a novel \textit{PhysAI-DS} dataset. From this dataset, we aim to address multi-hop inference in physics via a neuro-symbolic approach designed to account for both the linguistic and mathematical features of physics argumentation and explanation.

\bibliography{apssamp}

\end{document}